\pdfoutput=1
\documentclass{article}

\usepackage{setup}

\begin{document}



\def\TOPIC{Deep Learning}

\def\TITLE{\emph{Deep} Convolutional Neural Networks}
\def\SUBTITLE{A survey of the foundations, selected improvements, and some current applications}

\def\AUTHORS{
    Lars Ankile (larslank) \\
    Morgan Heggland (morganfh) \\
    Kjartan Krange (kjartkra) \\
}


\begin{titlepage}

\vbox{ }

\vbox{ }

\begin{center}
\includegraphics[width=0.40\textwidth]{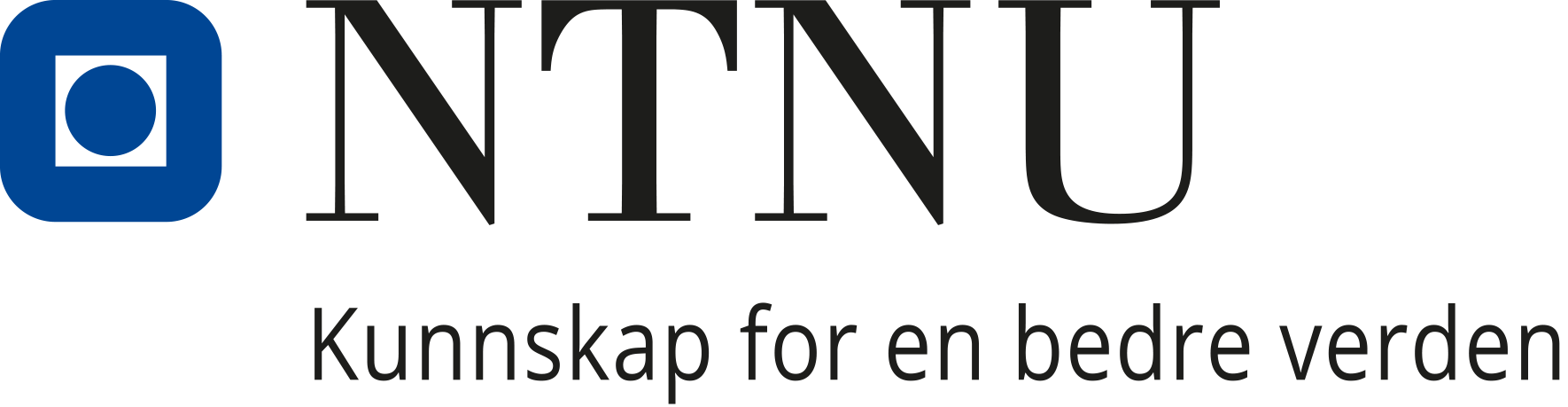}\\[1cm]
\textsc{\LARGE Department of Computer Science}\\[1.0cm]
\textsc{\Large TDT4173 - Method Paper}\\[0.5cm]

\vbox{ }
\HRule \\[0.4cm]
{ \large \TOPIC } \\[0.3cm]
{ \huge \bfseries \TITLE}\\[0.4cm]
{ \large \bfseries \SUBTITLE}\\[0.3cm]
\HRule \\[1.5cm]
\large

\emph{Authors:}\\
\AUTHORS

\vfill
{\large October 12, 2020}
\end{center}
\end{titlepage}






\frontmatter

\begin{abstract}
    Within the world of machine learning there exists a wide range of different methods with respective advantages and applications. This paper seeks to present and discuss one such method, namely Convolutional Neural Networks (CNNs). CNNs are deep neural networks that use a special linear operation called convolution. This operation represents a key and distinctive element of CNNs, and will therefore be the focus of this method paper. The discussion starts with the theoretical foundations that underlie convolutions and CNNs. Then, the discussion proceeds to discuss some improvements and augmentations that can be made to adapt the method to estimate a wider set of function classes. The paper mainly investigates two ways of improving the method: by using locally connected layers, which can make the network less invariant to translation, and tiled convolution, which allows for the learning of more complex invariances than standard convolution. Furthermore, the use of the Fast Fourier Transform can improve the computational efficiency of convolution. Subsequently, this paper discusses two applications of convolution that have proven to be very effective in practice. First, the YOLO architecture is a state of the art neural network for image object classification, which accurately predicts bounding boxes around objects in images. Second, tumor detection in mammography may be performed using CNNs, accomplishing 7.2\% higher specificity than actual doctors with only .3\% less sensitivity. Finally, the invention of technology that outperforms humans in different fields also raises certain ethical and regulatory questions that are briefly discussed.
\end{abstract}
\clearpage

\tableofcontents


\mainmatter

\section{Introduction}
\label{sec:introduction}
\begin{figure}[H] 
    \centering
    \includegraphics[width=0.6\textwidth]{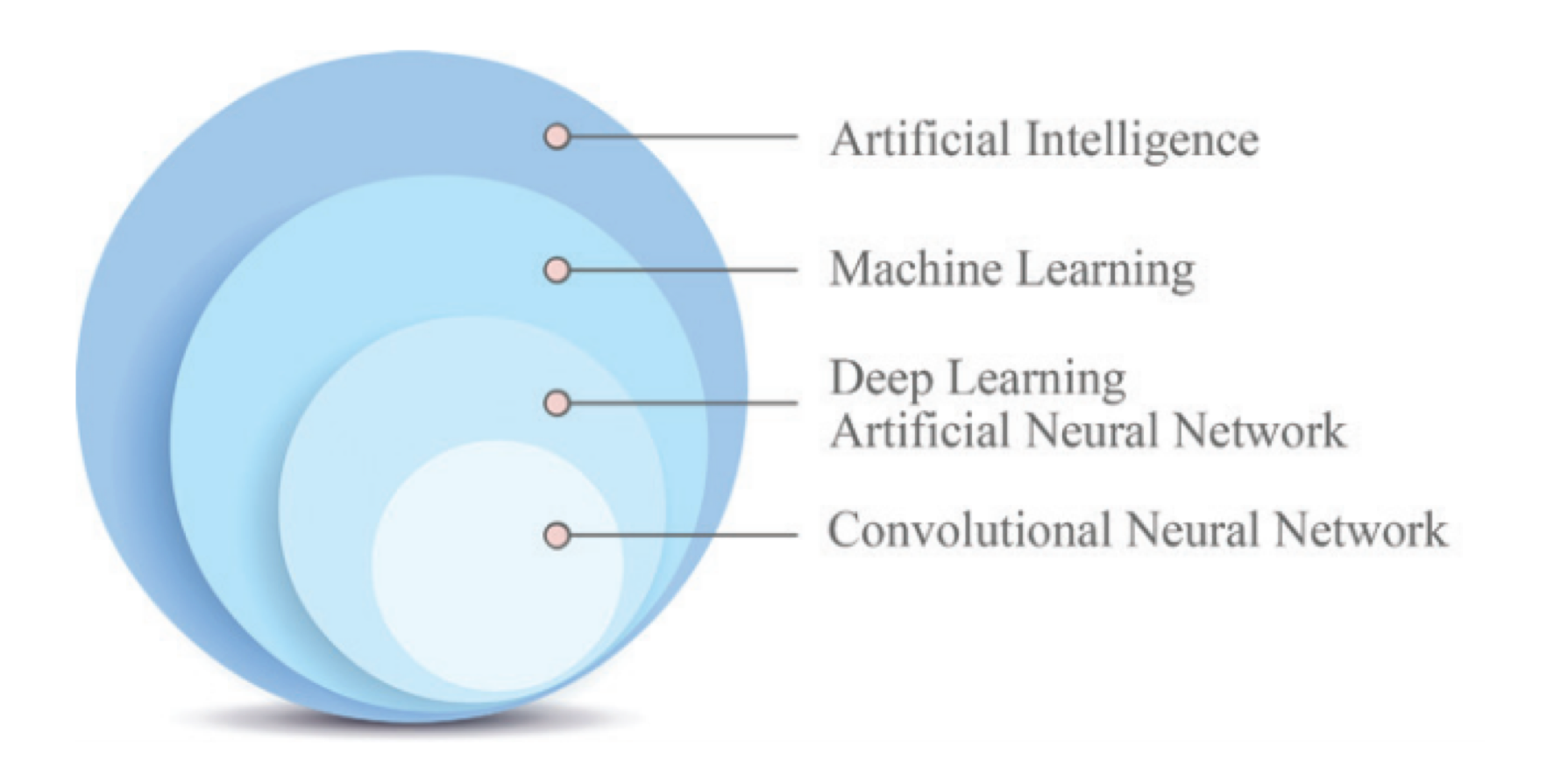}
    \caption{CNNs are a subset of Deep Learning Neural Networks.}
    \source{\cite{radiology}}
    \label{figure_1}
\end{figure}

In the  Machine Learning taxonomy, Convolutional neural networks (CNNs) are a subset of deep learning algorithms, as displayed in \autoref{figure_1}. CNNs can be both supervised and unsupervised. An example of the former is image classification and for the latter, an instance is word embedding (connecting words and their meanings).

CNNs are a type of neural network which are used to analyze data with a grid-like structure. A good example of this is image data, represented as two dimensions of RGB values. For problems such as image classification and the like, there are three main challenges with using a regular MLP (multi layer perceptron), which CNNs solve:

\begin{itemize}
    \item \emph{Parameter growth:} Using one perceptron for each pixel causes the amount of parameters to increase rapidly.  
    \item \emph{Translations:} A standard MLP would treat a picture and its slightly shifted version as two entirely different images. Recognizing a car in a picture should not depend on where in the image the car is located. 
    \item \emph{Spatiality:} MLPs do not account for spatial relations in images. The fact that two pixels are in proximity to one another is meaningful information.
\end{itemize}

Thus, CNNs solve the problem of understanding images, using networks of more manageable complexity. The special neural network takes into account that physical closeness between pixels bears meaning and that elements of interest can appear anywhere in a picture. This is accomplished through the use of a linear convolution operation, which is discussed in \autoref{sec:foundations}. The use of this operation, in one or more layers, is what defines a CNN.

Sometimes, however, the assumptions underlying the design choices of CNNs must be weakened or altered due to the nature of the input data. Translational invariance is, for example, not always as desirable. Furthermore, even though computational complexity is more manageable, modern networks tend to be very deep. This makes clever algorithms for calculating the convolutions a must. Both of these aspects are discussed in detail in \autoref{sec:improvements}.

Today, the method enjoys a wide range of applications, spanning from object recognition to natural language processing. CNNs like YOLO (\autoref{sec:applications}), are used to find objects and are used in e.g. BMW cars. In 2018, Cisco® prophesized that by 2022 84\% of all data traffic would be video \citep{cisco}. Even more striking, this high percentage is not counting still images. Since computers traditionally have had a notoriously hard time understanding visual data, the growing amount of videos and images on the world wide web makes the need for understanding this data apparent. The evolution CNNs have seen over the last decade is a huge step in the direction of making visual data readable by computers. In fact, convolutional neural networks have won every single ILSVRC (ImageNet Large Scale Visual Classification Challenge) since 2012 indicating that CNNs are the best publicly available solution for image classification problems.

\section{Foundations}
\label{sec:foundations}
The use of convolution instead of matrix multiplication in at least one of the layers in a neural network is what defines a convolutional neural network \citep{Goodfellow-et-al-2016}. "Use of convolution" refers to the \emph{convolutional operator}, a linear operator which is an essential distinctive element of a convolutional neural network, and will be the focus of the subsequent sections. In this section, the core idea and theoretical foundations of the convolutional operator are discussed.

\subsection{Core Idea of Convolution}
Prior knowledge of the structure or nature of the input data to a neural network can in some cases be exploited to improve the performance of the network. One such example is signal processing, which includes both temporal (e.g. sound) and spatial (e.g. 2D images) signals \citep{MIT-6-036}. This data can be thought of as grid-like: sound waves and other time-series data resemble one dimensional grids where each measurement is an entry, images resemble a two dimensional grid of pixels etc. This allows for optimization of analysis of such data in a neural network. Compared to a traditional network, runtime and space complexity may be improved, and the training data generally required for feasible results may be reduced.

\begin{figure}[H] 
    \centering
    \includegraphics[width=0.4\textwidth]{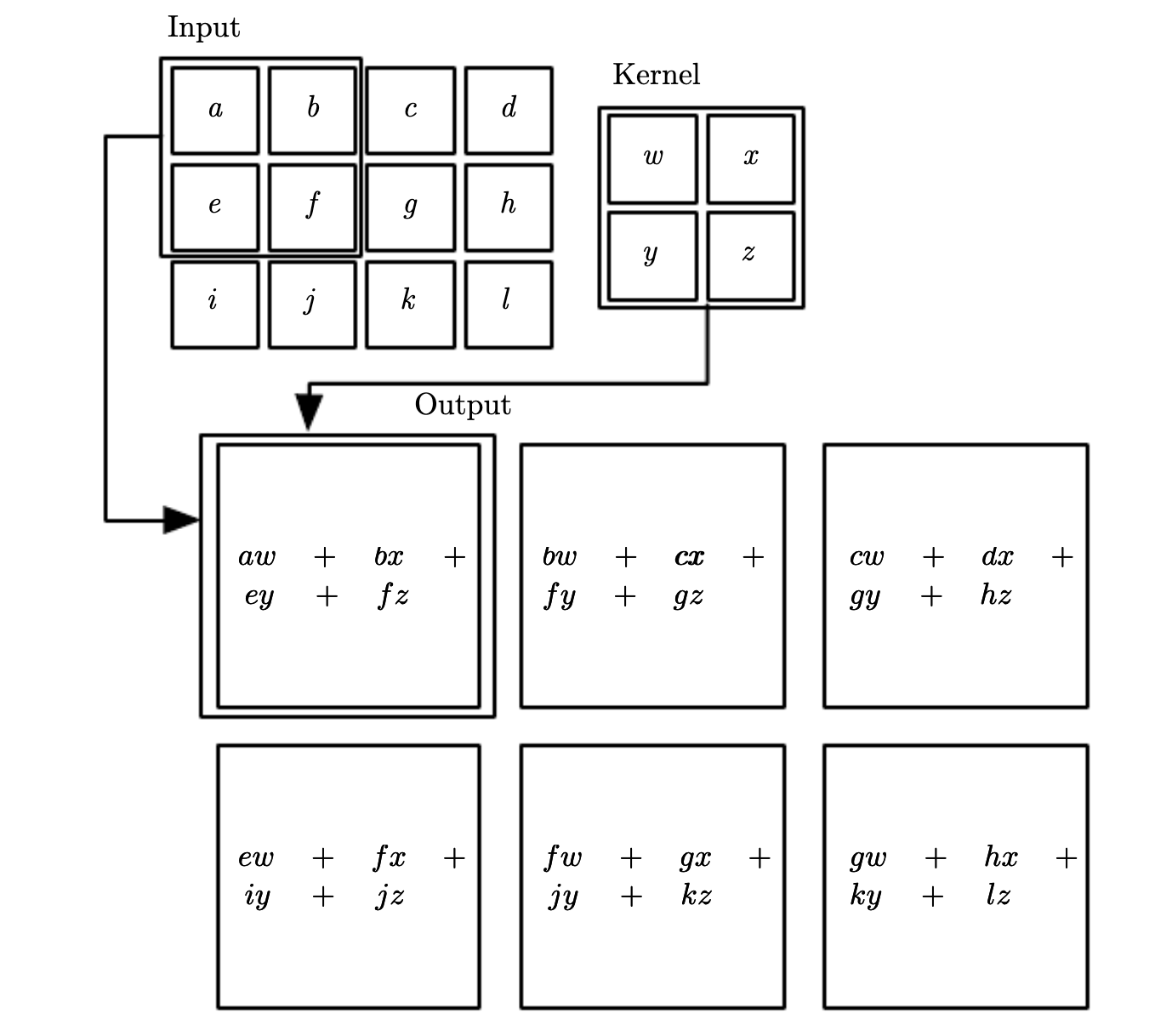}
    \caption{Example of convolution of two dimensional input data, ignoring outputs where the kernel is positioned partially outside the input data. The outlined output indicate the result from applying convolution with the given kernel on the outlined input.}
    \label{fig:2d-conv}
    \source{\cite{Goodfellow-et-al-2016}}
\end{figure}

The main idea behind convolution is the identification of features in the input data through the application of a \emph{kernel} (also known as \emph{filter}) across the input data. Both the input data and the kernel is of a grid-like structure, and can be represented as \emph{tensors}, which are multidimensional arrays. The kernel can be of any size, and is usually smaller than the input data. Kernels are used to identify features in the input data, such as edges in an image. The input data is \emph{convolved} with the kernel, meaning the kernel is "slid" across the input data, calculating the dot product or matrix product (depending on dimensions) between the overlapping part of the input data and the kernel. An illustrative example of the convolution operation can be seen in \autoref{fig:2d-conv} and a mathematical description can be found in \autoref{sec:theoretical_foundations}.

When applying convolution, it is common practice to add \emph{padding} to the edges of the input data to account for situations where the kernel is positioned partially outside the original input values. The padding is commonly implemented by extending the edges of the data with 0-valued entries (other values may be used). Furthermore, it is common to increase \emph{stride} in order to skip over indices when applying the kernel over the input, thus decreasing the size of the layer output.

There are a number of reasons that make this approach effective for this certain type of data. Firstly, it allows for the exploitation of a number of \emph{invariant} properties of the data. One such property is \emph{spatial locality}: the fact that the relevant information that makes up features are close to each other in a topological sense (e.g the pixels considered to find a cat's ears are close to each other spatially). Another property is \emph{translational invariance}: the fact that a feature can be identified independently of where it is located in the topology (e.g it a cat is a cat regardless of where it is located in an image). Both of these properties can be leveraged as a result of the nature of the convolutional operation: sliding the kernel over the input data to identify features. Informally speaking, the spatial locality is leveraged through applying kernels to an enclosed region of the input data each time, and translational invariance is leveraged through the movement of the kernel across the input data. Additionally, as a result of reusing the same kernel on the entirety of the input data, this method significantly reduces the number of parameters required to store and tune compared to traditional neural networks, which generally uses parameters only once. This concept is called \emph{parameter sharing} and is an essential advantage of CNNs. While traditional fully connected layers rely on a large number of parameters describing the interaction between each input and output unit, applying a kernel which is smaller sized than the input also significantly reduces the number of parameters required. This is called \emph{sparse connectivity}, as opposed to full connectivity. An illustration of this is shown in \autoref{fig:fc_conv_tiled_local}, and further discussion of these topics is included in \autoref{sec:improvements_to_the_basic_convolution}.

Normally in CNNs, the application of the convolution operation is followed by a \emph{non-linearity stage} and then a \emph{pooling stage}. This pipeline has proven to be an effective constellation for identifying features in the data \citep{Goodfellow-et-al-2016}. The main idea behind the non-linearity stage is the application of a non-linear function to the input data, ensuring the non-linearity properties of the network. The pooling stage further alters the output by using a summary operation on regions of the data, reducing the output in size while still highlighting key features, for instance by use of maximization aggregation (known as {\emph{max-pooling}}).

\subsection{Theoretical Foundation}
\label{sec:theoretical_foundations}
In a theoretical sense, the convolutional operator in its most primitive form can be considered an operation on two functions of a real-valued input \citep{Goodfellow-et-al-2016}. The convolution operation is defined as

\begin{equation}
s(t) =  (x*w)(t) = \int x(a)w(t-a)da
\end{equation}

where $x$ is the function mapping to a specific value in the input data, and $w$ represents the kernel. This formulation can for example (with the right kernel) be thought of as a smoothing average of $x$ across its entire domain, giving highest weight to values closest to $t$. If the input values are discrete, the convolution operation can be rewritten using summation:

\begin{equation}
    s(t) =  (x*w)(t) = \sum_{a=-\infty}^{\infty} x(a)w(t-a).
\end{equation}

The input is commonly multidimensional. In that case, one can replace the functions with multi-variable functions, operating on tensors. Consider an example of applying convolution to a two-dimensional image $I$ as input. One can then use a two-dimensional kernel $K$, and the operation can be written as follows:

\begin{equation}
    S(i,j) = (I*K)(i,j) = \sum_{m}\sum_{n}I(m,n)K(i-m,j-n).
\end{equation}

That is, for a given pixel in the input, positioned in row $i$ and column $j$, the convolution is computed by "placing" the centre of the kernel over the input pixel, and summing over the product of overlapping kernel parameters and input pixels to produce the output value for $i$ and $j$.

The above operation forms the basis for the operation performed in the convolutional stages or layers of neural networks. Depending on the values of the parameters of the kernels, this operation is able to highlight and identify different patterns and features in the input data.

\subsection{Training and Optimization}

The key component of convolution is the kernel with its corresponding parameters. The values of the parameters dictate which features and patterns the convolution operation identify. Thus, parameter selection is crucial in order to achieve high accuracy in solving the machine learning problem at hand. A method proven effective for setting these values is through training of the neural network in its entirety using labeled training data. As network size and complexity increase, fine-tuning a convolutional layer in isolation is infeasible, and performance is more easily measured by the output of the network as a whole. Only a brief overview of the training procedure is included in this subsection, as this is not the core focus of the paper.

\begin{figure}[H] 
    \centering
    \includegraphics[width=0.5\textwidth]{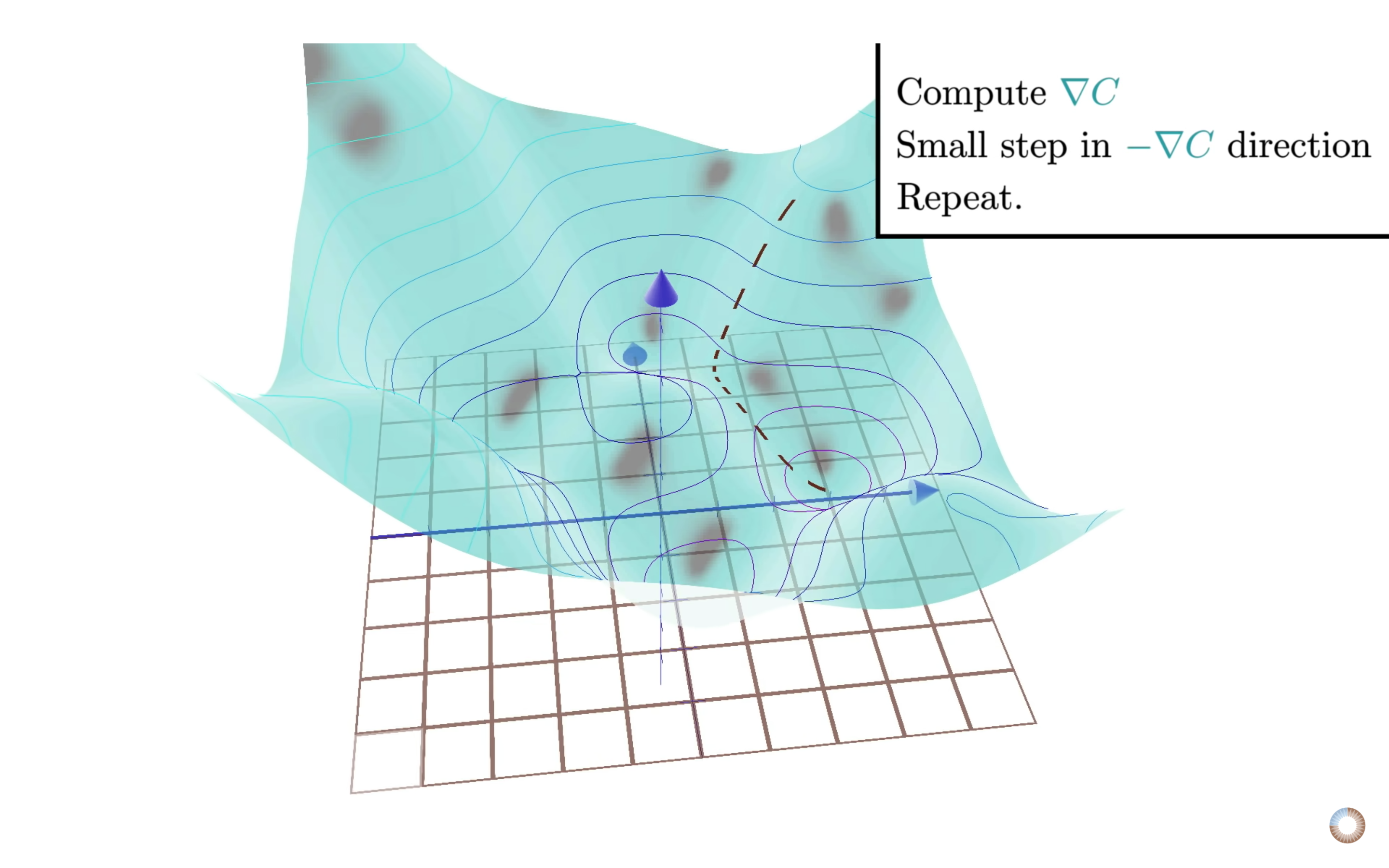}
    \caption{An illustration of gradient descent on a three-dimensional objective function.}
    \label{fig:gradient_descent}
    \source{\cite{3b1b}}
\end{figure}

Training is efficiently conducted using \emph{gradient descent} and \emph{back-propagation} \citep{MIT-6-036}. This method allows for automatic tuning of the parameters through the application of labeled training data. In essence, a \emph{loss function} or objective function calculates how well the network performed on a single input example compared to the true output (from the label). Based on this, one can calculate the gradient of the loss function evaluated at the training example with respect to the parameters or weights in each layer. This can be written as $\nabla_W Loss(NN(x:W), y)$, where $W$ represents all weights in all layers, $Loss$ is the loss function that takes a predicted value generated by the neural network $NN$ with input training example $x$ and weights $W$, and compares this to the actual value $y$. Using the chain rule of differentiation, this gradient can be decomposed into parts relevant to each layer. One can then apply the principle of back-propagation to calculate the gradient with respect to the weights in each layer. This represents the direction of the steepest ascent of the loss function with respect to each weight. By updating the weight in the negative direction of the gradient, taking a "step" according to a \emph{step size} parameter, the output loss is then reduced. This update can be written as $\theta_i := \theta_i - \eta\frac{\partial Loss}{\partial\theta_i}$, where $\theta_i$ is a parameter and $\eta$ is the step size. This is the principle of gradient descent and applies to kernel weights just as much as weights in a traditional neural network layer. An example of the descent can be seen in \autoref{fig:gradient_descent}.

As neural networks increase in size, the number of parameters grow significantly. In order to produce reliable predictions and avoid \emph{overfitting}, a large amount of diverse training data is required. Overfitting is when the product of an analysis fits its input data or training data "too well", in a sense that it will likely not generalize well to new data. By leveraging convolution, less data can be used without the risk of overfitting. This comes as a result of the benefits of convolution, such as parameter sharing and sparse connectivity. On the other hand, increased use of parameter sharing may cause \emph{underfitting}. That is, the network is not able to capture the full predictive power available in the training data. Alternatives to help alleviate this issue are discussed in \autoref{sec:improvements}.

\section{Improvements to the Method}
\label{sec:improvements}

\subsection{Improvements to the Basic Convolution}
\label{sec:improvements_to_the_basic_convolution}

Convolutions and pooling impose an infinitely strong \emph{prior} on the model parameters \citep{Goodfellow-et-al-2016}. A prior probability distribution over the model parameters is a way to encode our beliefs about what kind of models are likely to be good before the model is ever presented any data. The strength of the prior says something about how strong our belief is. A weak prior let the parameters roam relatively freely during training, while a strong prior restricts the parameters. An infinitely strong prior sets the probability for some of the parameters to be 0, i.e. they can never influence the model, regardless of how much support the data provides.

\begin{figure}[H]
    \centering
    \includegraphics[width=0.9\textwidth]{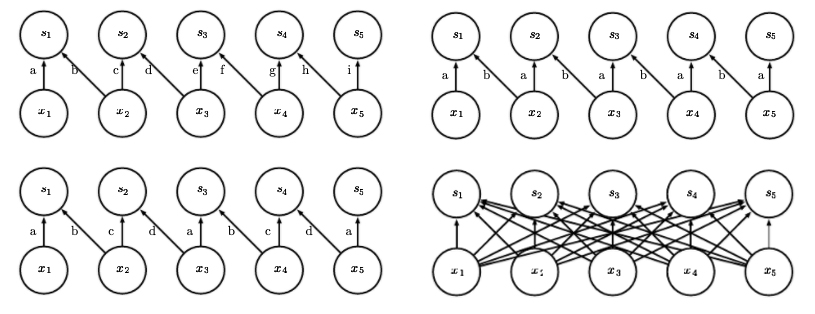}
    \caption{Comparison of the connectedness of convolutional and fully connected layers, as well as of degree of parameter sharing between different layer architectures. All but the network in the bottom-right (fully connected layer) are locally connected. The locally connected networks differ in the degree to which they share parameters. In the top-right network (standard convolution, kernel size 2), we see that the same two parameters are shared among all the connections. In the top-left network (locally connected layer), there are a separate parameters for every connection. In the bottom-left network (tiled convolutional layer, kernel size 2, 2 distinct kernels), there is some sharing, albeit with more separation and more parameters.}
    \label{fig:fc_conv_tiled_local}
    \source{\cite{Goodfellow-et-al-2016}}
\end{figure}

In the case of convolutional networks, the infinitely strong prior says that the function that the convolutional layer learns will only contain local interactions and is invariant to translation \citep{Goodfellow-et-al-2016}. A convolutional layer can be considered a fully connected layer where all non-local connections are set to have a permanent zero-weight (this is not how it is done in practice but is a useful conceptual tool). See \autoref{fig:fc_conv_tiled_local} for a visual comparison of convolutional layers and fully connected ones.

For any prior to be useful, we would require the assumptions made by the prior to be accurate. E.g. when inputs in distant locations influence the output, the prior imposed by the convolution might be too strong and cause the network to underfit. In this subsection, we will present two alternatives to the basic convolution that weaken this prior slightly, \emph{locally connected layers} and \emph{tiled convolution}.

\subsubsection{Locally Connected Layers}

As stated above, the convolutional layer is invariant to translations. Sometimes, however, approximate position information must be preserved. An example can be seen in \cite{LeCun1989BackpropagationAT}. One solution to the problem is to use locally connected layers. This form of layer preserves the local connections to take advantage of the fact that local interactions is what matters in images, but can treat different parts of the image differently. This is useful when we know that features will be a function of a local area of the input space, but have no reason to believe that we will find the same features multiple places in the input \citep{Goodfellow-et-al-2016}. An example could be in facial recognition, where the eyes usually would be located near the top of the image. Locally connected layers can also be beneficial when using very deep, zero-padded convolutions. In those networks, the inputs along the edges will have less and less impact on the output as we go deeper into the network. Because locally connected layers do not share parameters, they can learn to amplify the edge inputs more than the ones in the center, if useful given the training data.

In \autoref{fig:fc_conv_tiled_local}, one can see an example of a layer with local connections in the top row. Observe that each node is equally connected as in the convolutional case (bottom), but each connection has a unique parameter associated with it.

Locally connected layers will require orders of magnitude fewer parameters than the fully connected ones because of being locally connected from layer to layer. However, they will require many more parameters than regular convolutional layers because they do not utilize parameter sharing. This can lead to a large performance-hit in use-cases where the properties of locally connected layers are not required.

\subsubsection{Tiled Convolution}

Convolutional neural nets allow for hard coding of translational invariance into the network. However, this makes for a rigid structure that is unable to adapt to other kinds of invariances. Examples are invariance to scale and rotation, both of which regular convolutions handle poorly. \cite{NIPS2010_4136} proposes \textit{tiled convolutional neural nets}, or Tiled CNNs, as an alternative. The main benefit of this approach is that it allows the network to \textit{learn} invariances, instead of them having to be hardcoded into the network. This architecture can learn complex invariances (e.g. rotation and scale) while still keeping much of the advantages of CNNs' low number of parameters, which allows for faster training and easier learning (compared to fully connected layers).

What sets tiled CNNs apart is that they only constrain parameters that are $k$ steps away from each other. The special case of $k = 1$ yields the normal convolution operation, and with $k$ equal to the number of input parameters we get fully untied simple units (like locally connected layers discussed above). When we vary $k$ along this spectrum we vary the trade-off between having few learnable parameters and being able to learn complex invariances.

In \autoref{fig:fc_conv_tiled_local}, one can see a comparison between locally connected layers, tiled convolution, and standard convolution. In the example, the network in the middle is a simple representation of tiled convolution, where each kernel is 2 pixels wide, and there are 2 distinct kernels. This means that it is just as connected as the regular convolution (bottom), but there is twice the number of parameters, and the parameters that are forced to be equal are more spaced out.

\begin{figure}[H]
    \centering
    \includegraphics[width=0.8\textwidth]{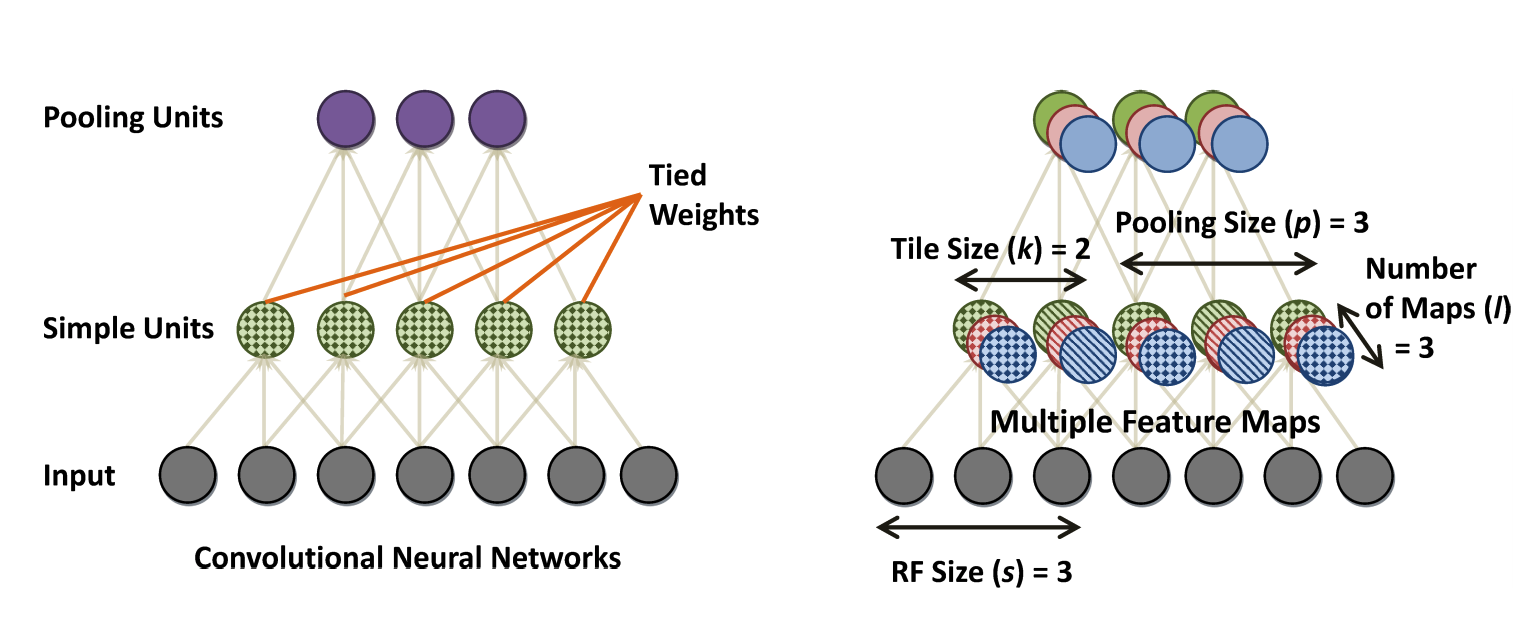}
    \caption{Left: CNNs with local connections and shared weights. Right:
Partially unshared locally connected networks – Tiled CNNs. Units with the same color belong to the
same feature map, and within each map, units with the same fill texture have tied weights.}
    \label{fig:tiled_conv}
    \source{\cite{NIPS2010_4136}}
\end{figure}

\subsection{Improvements to the Computation of the Convolution}

The deep CNNs of modern machine learning often contain millions of parameters, making the use of parallel processing essential to be able to train the networks in reasonable time \citep{Goodfellow-et-al-2016}. In some special cases, it is also possible to speed up the convolution by utilizing different convolution algorithms.

\subsubsection{Using the Fast Fourier Transform to Convolve, \textit{Fast}}

In \cite{Mathieu2014FastTO}, the \textit{Fast Fourier Transform} (FFT) is utilized to significantly increase efficiency of training and inference in CNNs. The work builds on the \textit{Convolution Theorem}, that states that convolutions in the space domain are equivalent to point-wise product in the Fourier domain. Let $\mathcal{F}$ be the Fourier transform and $\mathcal{F}^{-1}$ be the inverse. We have that the convolution of two arbitrary functions $f$ and $g$ is

\begin{align}
    \label{eq:fourier-conv}
    f * g = \mathcal{F}^{-1}(\mathcal{F}(f) \cdot \mathcal{F}(g)).
\end{align}

The increased efficiency is achieved by both utilizing that the dot product is a faster operation than the convolution operation, as well as that the transformed input features can be reused multiple times, lowering the initial cost of transforming them.

There are mainly 3 equations containing convolutions that are important for why this approach is efficient. Let $x_f$ be a set of input images indexed by $f$, of dimensions $n\times n$ and $y_{f'}$ the set of output images indexed by $f'$, and the layer $L$'s trainable parameters $w_{f'f}$, which are small kernels with dimensions $k\times k$, then the equations are the following:

\begin{align}
    y_{f'} &= \sum_{f} x_f * w_{f'f} \label{eq:conv-eq-1} \\
    \frac{\partial L}{\partial x_f} &= \frac{\partial L}{\partial y_{f'}} * w^T_{f'f} \label{eq:conv-eq-2} \\
    \frac{\partial L}{\partial w^T_{f'f}} &= \frac{\partial L}{\partial y_{f'}} * x_f. \label{eq:conv-eq-3} 
\end{align}

As mentioned, there are two things that make this approach fast. The first is that computing the convolution directly in general requires more operations than calculating the convolutions by going via the Fourier domain. In \cite{Mathieu2014FastTO} they compute that direct convolution and FFT convolution, respectively, for \autoref{eq:conv-eq-1} would require

\begin{align}
    & \textnormal{direct convolution:} & S\cdot f'\cdot f\cdot(n-k-1)^2\cdot k^2 \label{eq:direct-conv} \quad \textnormal{and} \\
    & \textnormal{convolution with FFT:} & 2Cn^2\log n [f'\cdot S + f\cdot S+f'\cdot f] + 4S\cdot f'\cdot f\cdot n^2, \label{eq:fft-conv}
\end{align}

operations respectively, with $S$ being mini-batch size and $C$ a constant. The expression for direct convolutions, \autoref{eq:direct-conv}, is a product of 5 variables, and will in general be larger than the expression for convolutions via FFT, \autoref{eq:fft-conv}, which is a product of 4 variables.

The second effect that provides the efficiency increase is that Fourier transforms can be calculated once and reused \citep{Mathieu2014FastTO}. That is because in equations \eqref{eq:conv-eq-1}, \eqref{eq:conv-eq-2}, and \eqref{eq:conv-eq-3}, all the matrices indexed by $f$ is convolved with each of the matrices indexed by $f'$. Therefore, each feature map can be processed by the FFT once, and all subsequent convolutions can be performed as pair-wise products in the Fourier domain, without having to recalculate the FFT. This makes for a large improvement in speed, and means that even though a single convolution might be more computationally expensive with FFT, the effective reuse of transformed inputs compensate for the overhead.
\section{Current Applications}
\label{sec:applications}
\subsection{YOLO}
YOLO, You Only Look Once, is one of the most efficient object detection algorithms that exist \citep{YOLOv4}.  Combined with some other clever approaches, it is an example of a CNN used to analyze visual data. The first version was published by \cite{Redmon2016YouOL}, with the newest implementation, YOLOv4 released in April 2020. For v4 and onward, Alexey Bochkovskiy is leading the project.

\begin{figure}[H] 
    \centering
    \includegraphics[width=0.5\textwidth]{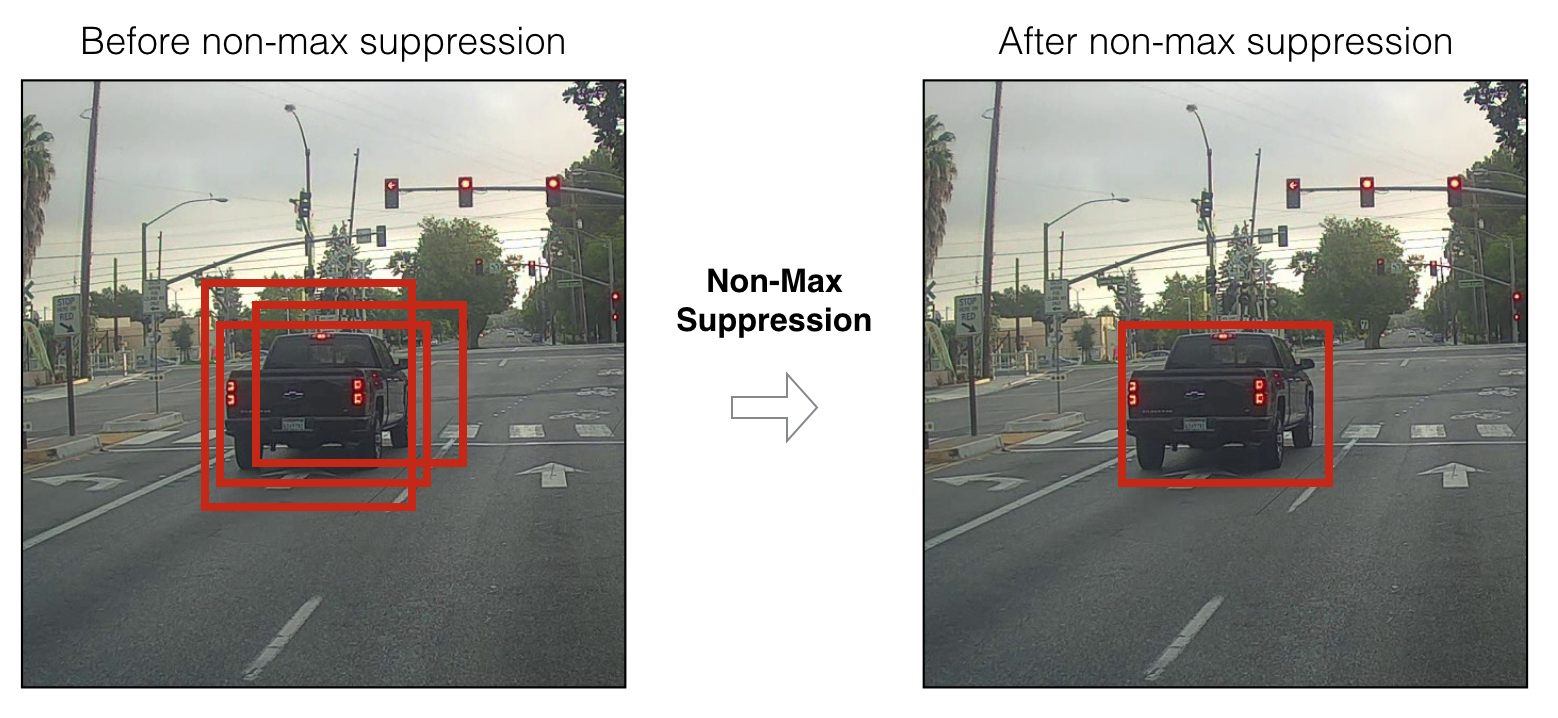}
    \caption{A subroutine of the YOLO algorithm – non-max suppression only keeps best fit boundaries }
    \source{\cite{Non_max_suppression}}
    \label{fig:non-max-suppression}
\end{figure}

The YOLO algorithm predicts both the position (represented as a bounding box) and the classification of objects in images. YOLO aims to find the following variables in a picture: $(bx,by)$ - the center of a bounding box, $(bw, bh)$ - the width and height of a bounding box, $c$ - the class of the object, $P_c$ - the probability that there is an object of class $c$ in the box. Often YOLO splits the image into a $19x19$ image, where each cell predicts $5$ bounding boxes of the form $y = (P_c,bx,by,bw,bh,c)$. This gives $19x19x5$ = $1805$ different bounding boxes per picture. Removing the boxes with a low $P_c$ is called \emph{non-max suppression} as seen in \autoref{fig:non-max-suppression}.

\begin{figure}[H]
    \centering
    \includegraphics[width=0.35\textwidth]{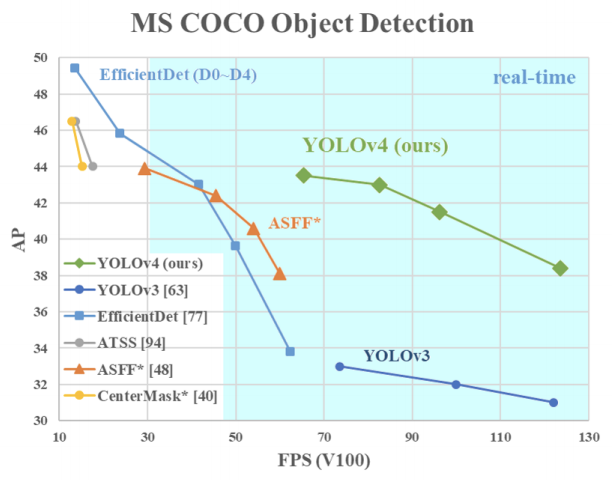}
    \caption{ Display of YOLOv4 vs. competing algorithms. v4 runs twice as fast as EfficientDet with around the same performance. AP stands for average precision and FPS stands for frames per second. }
    \source{\cite{YOLOv4}}
    \label{fig:yolov4}
\end{figure}

The main difference between YOLOv4 and earlier implementations is a focus on speed. The new YOLO algorithm's goal is that anyone having a decent GPU should be able to apply the algorithm to achieve incredible accuracy for image recognition running in real time. In \autoref{fig:yolov4}, the results from the comparison between YOLOv4 and other architectures is shown, where it is apparent that YOLOv4 outperforms most other neural networks in terms of average precision and does so at more than twice the frame rate. This differentiates YOLOv4 greatly from the other algorithms, as it can be used in real time classification of objects at a near human precision. For instance, the CNN can differentiate between cars, bikes and trucks driving on a highway. With its high speed and surprisingly good precision YOLO is widely adopted and as such a success.  

\subsection{Tumor Detection in Mammography}
One intriguing application of convolutional neural networks is within the field of radiology. Each year millions of women go through treatment for detection of cancers at an early stage through mammography. A paper published in Nature Scientific Journal \citep{tumor}, shows that CNNs can already detect cancers at an early stage with high accuracy. In fact, their algorithm is in many ways already outperforming US doctors. The use of CNNs in diagnosis can greatly reduce the cost for hospitals, making tests available to a wider public and at the same time increase diagnostic precision. 

\begin{figure}[H] 
    \centering
    \includegraphics[width=0.8\textwidth]{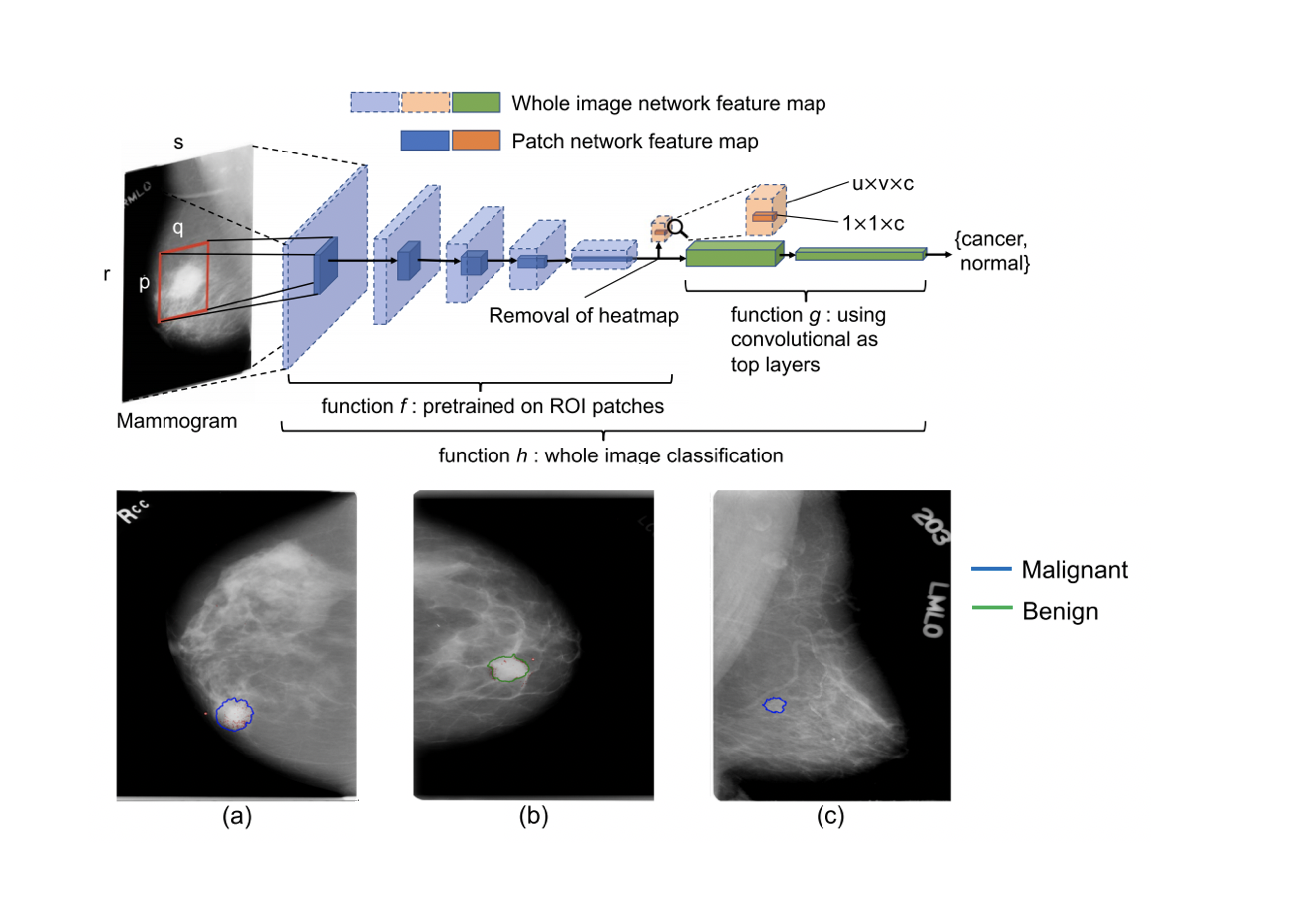}
    \caption{A depiction of the neural network architecture applied, and three mammography pictures, showing that the network quite clearly detects and diagnoses tumors.}
    \source{\cite{tumor}}
    \label{fig:tumor}
\end{figure}

The study presents a CNN that has the input of a mammographic picture and outputs a location and presence of benign (not dangerous) or malignant (should be treated) tumors. In medical diagnosis, \emph{sensitivity} refers to the ability of correctly identifying a sick person, the \emph{specificity} refers to identifying a healthy person correctly. In the US, the average sensitivity of digital mammography screening is 86.9\% with an average specificity of 88.9\%. As seen in \autoref{fig:tumor}, some of the tumors being spotted are quite hard for even a trained personnel to see.

The resulting CNN ran on multiple databases with an AUC (area under the ROC curve) on the lowest quality pictures of 0.91. Trained on full-field high resolution pictures of the INbreast database, their best model achieved a per-image AUC of 0.95. Combining all of their models, this number was raised to 0.98 with a sensitivity of 86.7\% and a specificity of 96.1\%! This shows that CNNs can perform tasks that are cost saving, but more importantly, lifesaving. With the result of the paper in mind, this would mean that the CNN is outperforming US medical personnel in detecting and classifying tumors. US doctors are only .3\% more likely to correctly spot a malignant tumor but have a 7.2\% less chance of correctly finding a patient to be healthy. This may imply that US doctors tend to diagnose malignant tumors when in doubt. As a last point, it is interesting to discuss the moral dilemma that arises from this study. It is important to keep in mind that the 7.2\% incorrectly diagnosed with a malignant tumor are unnecessarily placed under a lot of stress, chemotherapy and could potentially be scarred for life. Thus, the paper indirectly touches on the moral dilemma – should technology replace doctors as soon as it is better by some metric? As of March 2020, \cite{hospitals_applications} presents actual adoption of ML in hospitals as being in its infancy.
\section{Conclusions and Further Work} 
\label{sec:conclusions}

CNNs are deep neural nets which utilize the convolutional operation to materialize stunning results on problems of object detection and image classification. The convolutional part of CNNs are what sets them apart, and therefore the chosen focus of this paper. Furthermore, the paper laid out theory behind why convolution is thought to be effective in signal processing, and highlighted areas where improvements and optimizations are possible.

Of the many potential improvements, two have been discussed. First, locally connected layers which keeps the local connectivity of convolutional nets, but none of the parameter sharing. This results in a layer that can be adaptive to cases where certain features are only relevant in certain parts of an input map. Secondly, tiled convolution is in many ways a compromise between standard convolution and locally connected layers. Tiled convolution brings back some of the parameter sharing, but is still able to learn more complex invariances in the input than the regular convolution. Use of the Fast Fourier Transform can greatly improve computational efficiency in some cases because convolution is just pair-wise multiplication in the frequency domain, which is a ``cheaper'' operation. Also, many of the Fourier transforms can be computed once, and then reused multiple times, which further increases the performance gains attainable from using this approach.

Even though there are improvements to be made, applications of CNNs in practice have proven to be very effective. In the world of fast image recognition, the YOLO architecture is a state-of-the-art neural network which  accurately  predicts  bounding  boxes  around a variable number objects in near-real time. Lastly, since tumor detection in mammography can be performed using CNNs, and these are outperforming trained doctors by reporting fewer false positives, one should already be asking the question if CNNs should replace (or preferably aid) our medical experts. This demonstrates the high proficiency of CNNs today, and that even though improvements can be made, CNNs can already \emph{help} humanity solve some of its biggest challenges. In our opinion, it is not an all-or-nothing issue. Doctors are still immensely important, but technology can help them be more efficient by enabling them to aid a larger number of people in a shorter amount of time, and with higher quality. In our view, that is what Justin Bieber would call a \emph{``No Brainer''}, alternatively \emph{``Best of Both Worlds''}, as Hannah Montana would say.

The number of ways to apply CNNs seem to be endless, and new ideas and improvements surface constantly. One idea could be to use a CNN, like YOLO, to be an effective way to implement driving aids. YOLO's ability to detect objects like people or bicycles can be leveraged to increase safety on the roads. A second application could be to use CNNs to automatically categorize images (and the objects within them). Based on the output from the CNN, a company like Instagram, could automatically generate hashtags and categorize pictures of e.g. beaches or with dogs in them. The list of new applications could go on and on, but to mention a couple more: unlocking doors with facial recognition, automatic pet watching cameras, automatic pose detection in dance games, fraud detection (by treating transactions as a signal, as a function of time), and much more.

As a last note, we have some thoughts regarding the misuse of CNNs. As William Lamb, member of the British Parliament, said in 1817: \emph{``The possession of great power necessarily implies great responsibility.''} Joseph Redmon, the creator of YOLO, left the project because he did not want to work on something that he saw was used for military and aggressive marketing purposes. Another example is OpenAI's GPT language model (also utilizing deep neural nets, but not convolutions specifically). The model became capable of producing e.g. propaganda and hate speech at a much larger rate than humans, which could potentially be weaponized. The question Redmon asked himself is important to contemplate, even though one is easily awed and blinded by such brilliant technology. It is also important to remember that technology, which is basically a complex tool, is not evil, but it can be harnessed for evil purposes. In a field that is as fast moving as AI and ML, it is very hard for governments and regulatory agencies to be on top of all the developments, which makes appropriate regulation lag behind the curve.

\newpage
\bibliography{main}


\end{document}